\documentclass{article}

\usepackage{hyperref}  
\usepackage[accepted]{whi2016}
\icmltitlerunning{Visualizing textual models with in-text and word-as-pixel highlighting}
\usepackage{times}
\usepackage{graphicx} 
\usepackage{subfigure} 

\usepackage{natbib}

\usepackage{chngpage}
\usepackage{stmaryrd}

\usepackage{amssymb}
\usepackage{amsmath}
\usepackage{amsthm}
\usepackage{graphicx}
\usepackage{lscape}
\usepackage{subfigure}
\usepackage{chngpage}
\usepackage{latexsym}
\usepackage{nonfloat}

\newenvironment{itemizesquish}{\begin{list}{\labelitemi}{\setlength{\itemsep}{0em}\setlength{\labelwidth}{2em}\setlength{\leftmargin}{\labelwidth}\addtolength{\leftmargin}{\labelsep}}}{\end{list}}
\newcommand{\ignore}[1]{}

\usepackage{caption}

\begin{document}

\twocolumn[
\icmltitle{
Visualizing textual models with in-text and word-as-pixel highlighting
\\ \vspace{0.2in}
\includegraphics[width=7in]{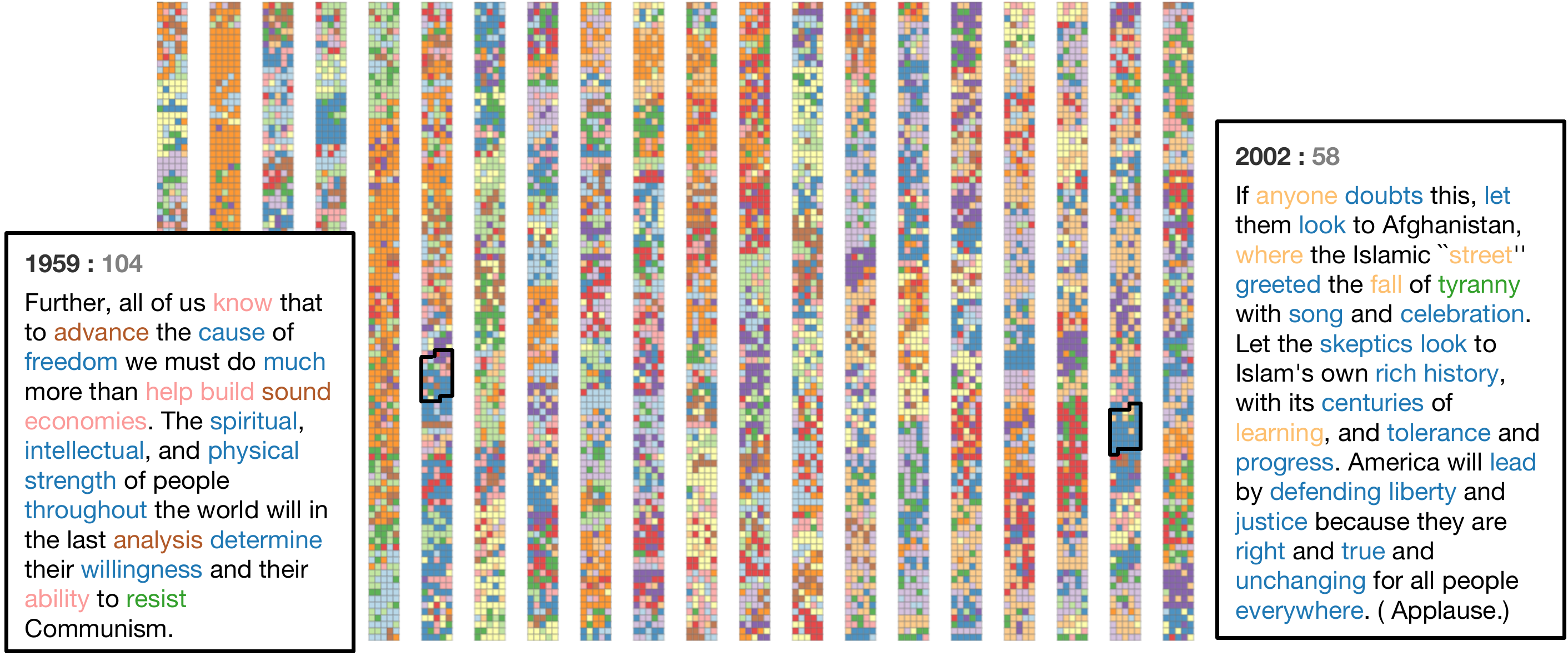}
\\ 
\captionof{figure}{A topic model's token-level posterior memberships $P(z_t|w_t)$ 
shown as
in-text annotation (\S\ref{s:intext})
and
word-as-pixel (\S\ref{s:wap})
views, from a corpus of U.S.\ presidential State of the Union speeches. 
Speeches are concatenated, running in columns; top-left is 1946, bottom right is 2007.
(This version shows a sample of tokens.)
\\ Demo: \url{http://slanglab.cs.umass.edu/topic-animator/}
\label{f:ldabig}}
}

\icmlauthor{Abram Handler}{ahandler@cs.umass.edu}
\icmlauthor{Su Lin Blodgett}{blodgett@cs.umass.edu}
\icmlauthor{Brendan O'Connor}{brenocon@cs.umass.edu}
\icmladdress{
College of Information and Computer Sciences, University of Massachusetts, Amherst, MA, 01060, USA
}

\icmlkeywords{}

\vskip 0.3in
]


\begin{abstract}
We explore two techniques which use color to make sense of statistical text models.
One method uses in-text annotations to illustrate a model's view of particular tokens in particular documents. Another uses a high-level, ``words-as-pixels'' graphic to display an entire corpus.
Together, these methods offer both zoomed-in and zoomed-out perspectives into a model's understanding of text. We show how these interconnected methods help diagnose a classifier's poor performance on Twitter slang, and make sense of a topic model on historical political texts.
\vspace{-0.2in}
\end{abstract} 

\section{Introduction}

Probabilistic models of text are a core technology for natural language processing. Such models typically link words or phrases with semantic categories, like classes or topics.
When we analyze data with these models, we need to
understand how the method interprets text in order to perform
(1) exploratory and confirmatory data analysis and
(2) error analysis for engineering improvements. \\

Previous work on interpreting and understanding text models has focused on summarizing text at the semantic or category level---for instance, by showing a list of most probable words in a particular latent topic \citep{Gardner2010Viz,Chaney2012Viz}.

In this work, we emphasize that text is originally a sequence of symbols (characters or words),
intended for a person to \emph{read}.
A system can provide insight into what a text model is thinking by showing a user the original text with automatic \emph{in-text annotations} describing the model's inferences (\S\ref{s:intext}).
Such annotations can be shown abstractly with a zoomed-out \emph{words-as-pixels} view (\S\ref{s:wap}) of text.
We demonstrate our methods using
topic models on political speeches and language classification on dialectal Twitter.

%

\section{Models}  \label{s:models}

For all models that we consider, a document $d$ consists of a sequence of symbols 
$\{w_t: t=1..N_d\}$.  This could be a sequence of words, or a sequence of characters;
we refer to elements in such a sequence
as \emph{tokens} (though \S\ref{s:langid} examines a character-based model).
For a particular model, we define a token-level \emph{visual quantity of interest} $\psi_t$
for position $t$,
which corresponds to an interesting value in the model. 
 These $\psi_t$ values are then encoded as visual attributes when displaying the original token sequence directly to the user (\S\ref{s:intext}).

\subsection{Token-level models (LDA)} \label{s:lda}

First we consider models that define latent variables at the token level.
For example, the latent Dirichlet allocation \cite{Blei2003LDA} model of text
posits a document $d$ arises from a $\theta_d$ weighting over $K$ topics,
where each token has a latent class $z_t$, indexing which word distribution $\phi_k$ is used to generate word $w_t$: $P(w_t,z_t\mid\theta_d,\phi) = \phi_{z_t,w_t} \theta_{d,z_t}$.  
We conventionally describe $\phi_k$ as a \emph{topic}.

At a single token position, the posterior topic membership breaks down as a compromise between document prevalence versus lexical probability; LDA is able to learn interesting representations since individual documents tend to be about a subset of topics and individual topics tend to include a subset of the vocabulary. The probability of a given latent topic is:
\[ P(z_t = k \mid w_t,\theta,\phi) \propto P(z_t \mid \theta_d) P(w_t \mid \phi_{z_t}) \]
We consider the vector of membership probabilities to be the visual quantity of interest,
defining:
\[ \psi_t = [ P(z_t=k \mid w_t,\theta, \phi) ]_{k=1..K} \]

Although we demonstrate our method using LDA, the same approach and methodology would apply to other common text models. For example, supervised sequence models (such as conditional random fields; \citet{Lafferty2001CRF}) also place tokens into semantic categories using token-level variables which can be visualized, as is often done in annotations interfaces for information extraction.\footnote{e.g.\ Brat: \url{http://brat.nlplab.org/}}
Similarly, \citet{Karpathy2015Viz} give an excellent demonstration of visualizing latent states
of a character-level long short-term memory (LSTM) recurrent neural network using token-level, in-text annotations (like \S\ref{s:intext}) to help understand a machine learned model.


\subsection{Token-level posterior impacts (MNB, LogReg)}  \label{s:mnb}

Many models 
do not directly define random variables at the token-level,
but sufficient statistics
resulting from individual tokens have a clear interpretation
in terms of how they affect inferences on model variables.
An example is document classification, where the frequencies of words
impact the posterior probability of the document class.  

Concretely, we consider
multinomial naive Bayes \cite{McCallum1998NB},
whose generative assumption posits that each document $d$ has a discrete label $y_d$ 
(drawn from distribution $\pi$), and the document's tokens are independently
generated from a single topic $\phi_{y_d}$.

Given learned parameters $\pi$ and $\phi$, to classify a document, we utilize the posterior $P(y_d \mid w_1..w_{N_d}) \propto P(y_d) \prod_t P(w_t \mid y_d)$ and calculate the posterior log-odds between classes $a$ and $b$:
\[ \log \frac{P(y=a \mid \vec{w})}{P(y=b \mid \vec{w})}
= \log \frac{\pi_a}{\pi_b} + \sum_t 
\underbrace{\log \frac{P(w_t \mid y=a)}{P(w_t \mid y=b)}}_{\psi_t}
\]
We restrict our attention to comparing the model's relative preferences for two classes $a$ and $b$,
and define $\psi_t$ to denote the token-level logit weight for one token instance $t$ in the text,
representing how much that token contributes to the posterior prediction of the document class.  

A wide variety of other models in the supervised setting may also define $\psi_t$ terms; for example, logistic regression has a very similar form \cite{Ng2002NBLR}.
In the binary classification case,
with bias term $\beta_{0}$ and word weights $\beta_1..\beta_V$,
logistic regression can be formulated similarly as MNB
in the case where features are word counts $n_v=\sum_t 1\{w_t=v\}$ and 
the ``posterior'' log odds is 
$\log[p(y=a|\vec{n})/p(y=b|\vec{n})] = \beta_0 + \sum_v \beta_v n_v$,
in which case the token-level logits\footnote{One issue is that non-linear transforms of the word counts,
such as thresholding or log scaling, often improve classification performance \cite{Yogatama2015Features};
unfortunately, they 
do not correspond to a uniform per-token  impact.}
 are $\psi_t=\beta_{w_t}$.

In practice, for both LDA and MNB, the full generative model is rarely used for all the text;
for example, at the very least, terms are excluded due to being stopwords, punctuation,
having a very high or very low frequency
(e.g.\ \citet{BoydGraber2014Care})
or are filtered out during feature selection.
This causes many tokens to not be accounted for in the model and thus do not change the posterior.
For MNB, we define $\psi_t=0$ in such cases.

\subsection{N-gram features} \label{s:ngrams}

It is useful to define features over n-grams, where each instance comes from a \emph{span} in the text in the form of a [start position, end position) pair; e.g.\ span $[3:5)$ corresponds to a bigram $w_{3:5}$ in positions 3 and 4.  Using n-gram features, MNB is no longer a proper generative model of the text sequence $\vec{w}$
but is still widely used in this setting where the document's log-probability $\log P(\vec{w}|y_d)$ is
defined as
the sum over all n-grams in the model.
In this case we define the span-level weight $\psi_{s:e}$ in a similar manner as \S\ref{s:mnb}:
\[ \log \frac{P(y=a \mid \vec{w})}{P(y=b \mid \vec{w})}
= \log \frac{\pi_a}{\pi_b} + \sum_{s:e}
\underbrace{\log \frac{P(w_{s:e} \mid y=a)}{P(w_{s:e} \mid y=b)}}_{\psi_{s:e}}
\]
Since a single token may participate in multiple overlapping n-grams,
we define the token-level weight as the sum of the weights of all (overlapping)
n-grams
that include $t$:
\[ \psi_t = \sum_{(s,e): \ t \in [s,e)} \psi_{s:e}\]
This can be extended to logistic regression or other feature-based classifiers as well.
$\psi_t$ answers part of the counterfactual:
if the token $t$ was deleted,
the prediction's logodds would change\footnote{
This analysis ignores the impact of new n-gram features, bridging position $t$, 
that would be introduced;
on the other hand, the new text likely would not be a valid or likely text,
so perhaps the counterfactual viewpoint is limited.}
by $-\psi_t$.

Other linguistic features could also be visualized using color annotations.
For example, a syntactic dependency path $P$ is
a sequence alternating between tokens and directed edges
(e.g.~\citet{Mintz2009Distant}).
Unlike an n-gram, the set of tokens in a path is not necessarily contiguous.
But tokens can be colored in the same way through a 
$\psi_t = \sum_{P: t \in S(P)} \psi_P$ value:
for a word token, the sum of the model's weights for paths whose token set $S$ includes the token.
If dependency edges are shown alongside the text, they could colored in a similar way.


\section{In-text visualization}  \label{s:intext}


We define a \emph{visual encoding function} $f(\psi_t)$ to select the final visual attributes to represent the quantity of interest to the user,
 inspired by \citet{Wilkinson2005GG}'s grammar of graphics approach to data visualization.
 

We would like to show the original text, with visual annotations.
Some easily implementable
options for visual encoding include \vspace{-1em}
\begin{itemizesquish}
\item Color: the background or foreground text color.
\item Boldface or italics.
\item Underlines (possibly varying color or line width).
\item Size of text.
\end{itemizesquish}\vspace{-1em}
In our preliminary experiments, color emerged as an effective encoding scheme.
Color can represent multiple dimensions as well as scalar values.
Previous research in visualization has examined 
how to effectively encode data in color
given the strengths and weaknesses of the human visual system
(e.g.\ \citet{Ware2012Book,Munzner2014Book}),
and research results such as
the ColorBrewer palettes\footnote{\url{http://colorbrewer2.org/}, \url{https://bl.ocks.org/mbostock/5577023}}
are available for use.
(On the downside, colors can pose an issue for colorblind users.)
Text size is another interesting option,\footnote{
Both word and tag clouds have long sought to encode frequencies from a bag-of-words using text size, e.g. the Wordle system (\url{http://www.wordle.net/})} --- 
but unfortunately variable sized text is often difficult to read.  Boldface and italics have a relatively limited information capacity, and we found underlines visually busy.
(An alternate approach is to use extra-textual visual cues alongside words; for example,
\citet{Chahuneau2012WordSalad} aligns a bar graph (heights corresponding to $\psi_t$)
next to word tokens.)

For the vector-valued $\psi_t$ from LDA, we assign different topics to different color hues (but similar brightness levels)
and assign a token's color according to the argmax of $\psi_t$.  (An additional possible strategy may be to blend the color towards white if the posterior entropy is higher.)

For binary document classification with a scalar-valued $\psi_t$,
there is a \emph{diverging} semantics:
negative $\psi$ and positive $\psi$ should correspond to different colors (e.g.\ red versus blue),
blending to white at $\psi=0$.
We utilize this for classifier visualizations.

\section{Words-as-pixels visualization}  \label{s:wap}

Color can be used in zoomed-out views as well.
For a very high aggregation level, such as
summarizing topic frequencies across thousands of documents over time,
the same colors can be used as the in-text annotations to assist interpretation.

We propose a complementary, high-level level view---\emph{words as pixels},
shown in Figure\ \ref{f:ldabig}.
Here, individual tokens are represented as pixels or very small squares with
coloring from their $\psi_t$, and these points are laid out in order within a document. We arrange as left-to-right descending columns, mimicking the natural reading order 
of many left-to-right languages, and thus corresponding to a zoomed-out view of the original text.\footnote{An inspiration is the zoomed-out scrollbar view of the Sublime Text editor.}

When documents have a natural ordering, such as date of publication, or sections or chapters within a book, multiple documents can laid out one after another. This allows the user to see certain discourse structures in the text; at least, ones that are captured by the model.
In Figure \ref{f:ldabig}, we visualize LDA on a corpus of U.S.\ presidential 
State of the Union speeches from 1946--2007 
using David Mimno's \emph{jsLDA} data preprocessing and topic model implementation
(\citeyear{Mimno2016JSLDADraft}).\footnote{\url{https://github.com/mimno/jsLDA} \ \ 
The words-as-pixels view only shows tokens that are in the model,
which here is roughly half of all tokens in the text after preprocessing.}
We average 100 Gibbs samples to calculate the $P(z_t|w_t)$ posteriors to be the $\psi_t$ quantities.
The model clearly picks up on natural local groupings of latent topics in the text.  
This is driven in part by the model assumptions encoded in data preprocessing,
since this version of the corpus defines model ``documents'' as paragraphs from the speeches.  The model assumption is that topic prevalence can be expected to vary by textual locality, and the visualization allows a qualitative assessment of to what extent this assumption 
holds in the posterior inferences.

This is apparent in the example: for example, large streaks of orange correspond to detailed discussions
of budgets that were common in the 1940s and 1950s.
We include callouts of individual paragraphs with a strong blue topic prevalence: discussion of political ideologies with regards to Communism (Eisenhower in 1959) and Islam (Bush in 2002).
We aim to develop this interface
as a \emph{linked views} data explorer \cite{Buja1996Interactive,OConnor2014MTE}
where a user can click on the word-as-pixel view to show the corresponding text passage.
A web demo is available at
\url{http://slanglab.cs.umass.edu/topic-animator/}.

\section{Classification: Language identification in social media} \label{s:langid}



A key step in any internet text analysis pipeline
is to identify which language a text was written in.
Character n-gram models (where each $w_t$ is a character symbol)
are a widely used approach for this task,
and the popular open-source \emph{langid.py} tool
\citep{Lui2011Langid,Lui2012Langid}\footnote{\url{https://github.com/saffsd/langid.py}}
uses a multinomial Naive Bayes model.

Short texts pose a challenge for language identification --- and
social media messages also present a domain adaptation problem,
since they contain much creative and non-standard language very different from traditional well-edited corpora that NLP systems are typically trained on. For example, \emph{langid.py} uses Wikipedia corpora as a major source of training data.

\begin{figure}
Predicted as Portugese (pt) \\
\hspace{-0.535in}
\fbox{\includegraphics[scale=0.55]{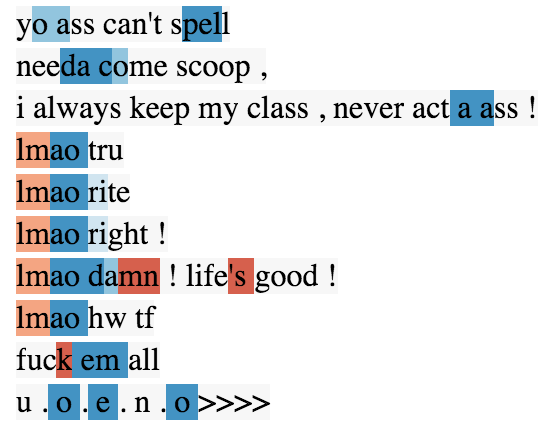}}
\\ \vspace{0.15in}
Predicted as Irish Gaelic (ga) \\ 
\fbox{\includegraphics[scale=0.55]{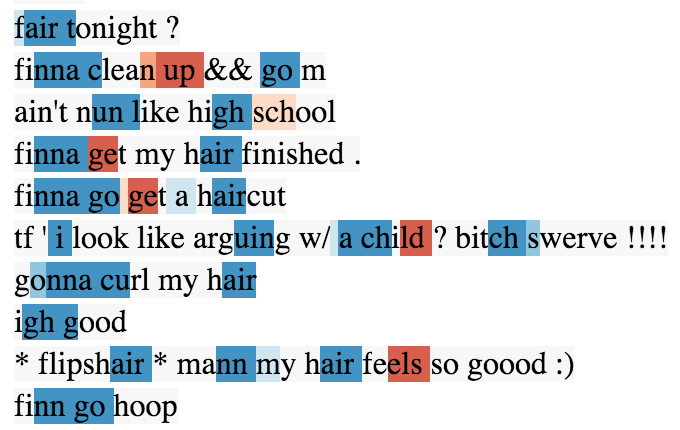}}
\caption{Tweets we assess as English that were classified as non-English;
every character position $t$ has is own $\psi_t$.
Blue indicates a $\psi_t$ log-likelihood weight towards the non-English language; red towards English.
\vspace{-0.2in}
\label{f:langid}}
\end{figure}

We examined a corpus of tens of millions public Twitter messages geolocated in the U.S., filtered to users who use language statistically associated with neighborhoods containing high populations of African-Americans.\footnote{Details in paper under review.}
As expected from the emerging sociolinguistic literature on social media corpora \cite{eisenstein2015review,eisenstein2015systematic, jones2015toward,  Jorgensen2015AAE}, these messages contain rich dialectical language very different from well-edited genres of English.
In fact, even after filtering only to messages only containing Latin-1 characters,\footnote{This filter gives the classifier an easier dataset more similar to its training data; for example, this  excludes emoji.}
\emph{langid.py} classifies 17\% of these users' messages as non-English, but upon inspection, nearly all of them are English.

We used in-text highlighting to help diagnose model errors (Figure~\ref{f:langid}),
assigning each character at position $t$ a color reflecting $\psi_t$,
the sum of all n-gram feature weights that fire at that position (\S\ref{s:ngrams}).

For example, the common term \emph{lmao} (\emph{laughing my ass off}),
ends in \emph{ao}, a common suffix in Portugese identified by the classifier.
The characters \emph{nna}, which are common in modal verbs in
non-formal American English (e.g.\ \emph{gonna}, \emph{wanna}, 
and the 
African-American English-associated \emph{finna} short for \emph{fixing to}) cause confusion towards Irish Gaelic.

Another result that we did not expect is the issue of sparsity in short texts.  Many messages have only a small number of firing features (which we anticipate could lead to low accuracy),
which is partly due to the feature selection process used to train \emph{langid.py}'s models, suggesting that its sparsity level may be tuned to a level more appropriate for longer documents than for these short ones.


%
%
%
%
%
%

\section{Conclusion}

This work stems from a fundamental aspect of text processing: the most natural and intuitive way to grasp the full meaning of a written text is simply to read it.
We believe in-text annotation is a less explored, but natural choice for explaining and \emph{understanding} a computer's view of language. We present a few simple methods for viewing text models,
but expect many avenues for future work.

\bibliography{example_paper,sulin}
\bibliographystyle{icml2016}

\end{document}